\documentclass[conference]{IEEEtran}
\IEEEoverridecommandlockouts
\usepackage{cite}
\usepackage{amsmath,amssymb,amsfonts}
\usepackage{algorithmic}
\usepackage{graphicx}
\usepackage{textcomp}
\usepackage{xcolor}
\usepackage{adjustbox}
\usepackage{hyperref}
\usepackage[font=small,labelfont=bf,tableposition=top]{caption}
\def\BibTeX{{\rm B\kern-.05em{\sc i\kern-.025em b}\kern-.08em
    T\kern-.1667em\lower.7ex\hbox{E}\kern-.125emX}}
\begin{document}

\title{FashionSD-X: Multimodal Fashion Garment Synthesis using Latent Diffusion\\
}

\author{\IEEEauthorblockN{Abhishek Kumar Singh}
\IEEEauthorblockA{\textit{ec22395@qmul.ac.uk} \\
\textit{Queen Mary University of London}
}}
\maketitle

\begin{figure*}[h!]
    \centering
    \includegraphics[width=1\linewidth]{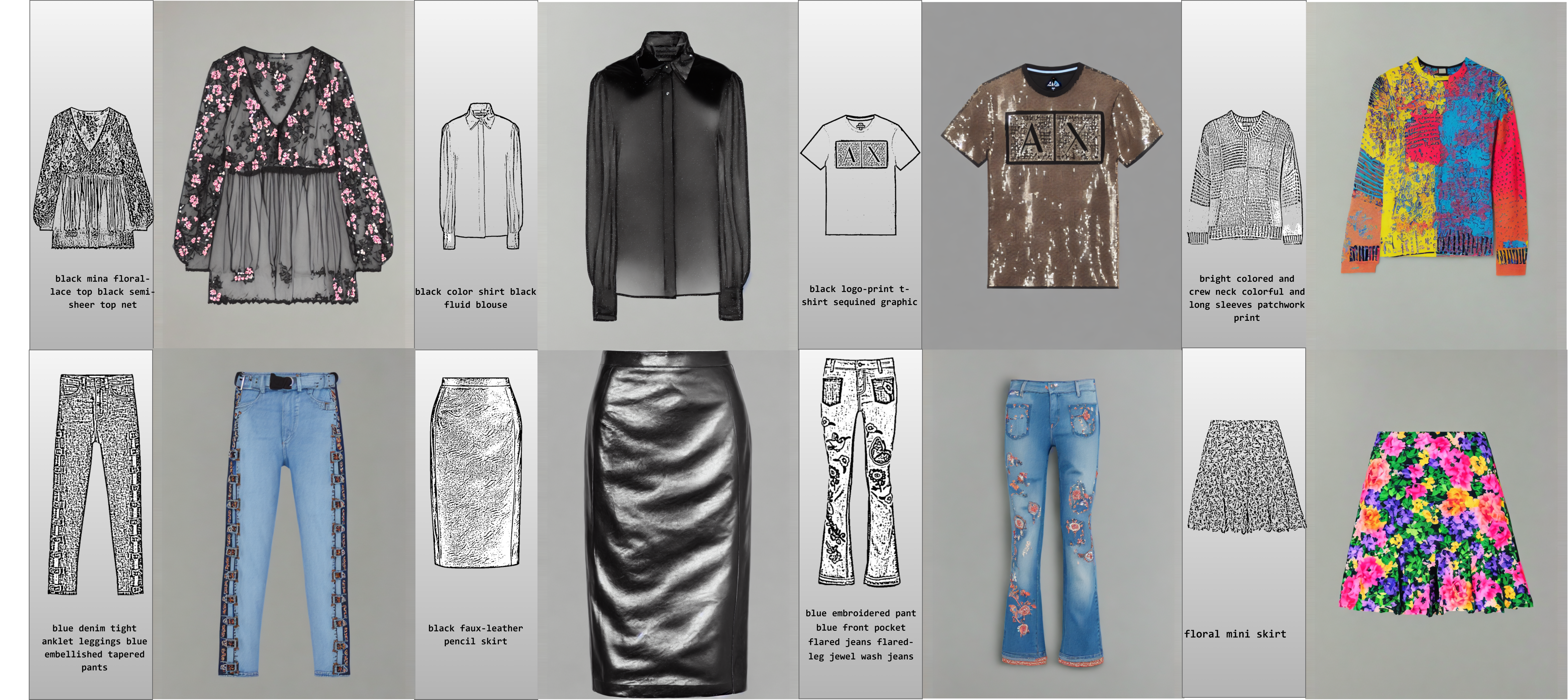}
    \captionof{figure}{Results of our proposed pipeline FashionSD-X, where the results are guided by a given input text prompt and sketch.}
    \label{fig:showcase}
\end{figure*}

\begin{abstract}
The fashion industry, in its constant quest for innovation, needs the integration of generative AI to help in the design process. We propose a novel pipeline to help fashion designers materialize their ideas, using latent diffusion models. In this paper, we introduce a pipeline using ControlNet and LoRA Fine-tuning that can generate high-quality images conditioned on multimodal inputs like text and sketch. We used state-of-the-art fashion virtual try-on datasets,  Multimodal Dress Code, and VITON-HD and extended them further to incorporate sketches. Our findings upon qualitative and quantitative evaluation based on, FID, CLIP Score, and KID showcase our model's superior performance compared to the vanilla stable diffusion model, signifying the potential for diffusion models to be involved in the fashion design process. This study paves the way for a more interactive, personalized, and technologically advanced future in fashion design and representation.

\end{abstract}


\section{Introduction}
The field of generative AI has experienced a considerable upswing of innovation and success over the past decade, introducing new frontiers in various domains, like art, design, and fashion. These tremendous developments are possible thanks to the emergence of powerful models capable of synthesizing high-quality images from
multimodal input. Researchers have started focusing on utilizing these models in the fashion domain, especially working on similar garment recommendation systems and virtual try-on clothes and accessories. But in this context, there haven’t been many significant works to help fashion designers generate clothing garments given a text description and a sketch and texture of the desired garment.\\

Diffusion models are in the spotlight due to their exceptional generative capabilities and the ability to cater to several downstream tasks in multiple domains. They define a Markov chain of diffusion steps that progressively add random noise to data and then learn to reverse this process, constructing desired data samples from the noise. Several notable diffusion-based generative models underlie these principles, including diffusion probabilistic models \cite{b3}, noise-conditioned score network \cite{b4}, and denoising diffusion probabilistic models \cite{b5}. A new kind of diffusion model known as the Latent Diffusion Models (LDMs) has emerged recently and has instantly become one of the leading diffusion models because of its low computational cost. The LDMs\cite{b6} work in the latent space of the pre-trained autoencoder, finding an optimal trade-off between computational load and image quality.  These models introduce a latent variable into the generative process, which provides a more expressive and flexible model of the data distribution. The latent variable allows the model to capture complex, multi-modal distributions and can be manipulated to control the generation process. Given their adaptability and outstanding generation capabilities, the use of diffusion models in the fashion domain is still only being remotely explored.

\begin{figure*}
    \centering
    \includegraphics[width=1.0\textwidth]{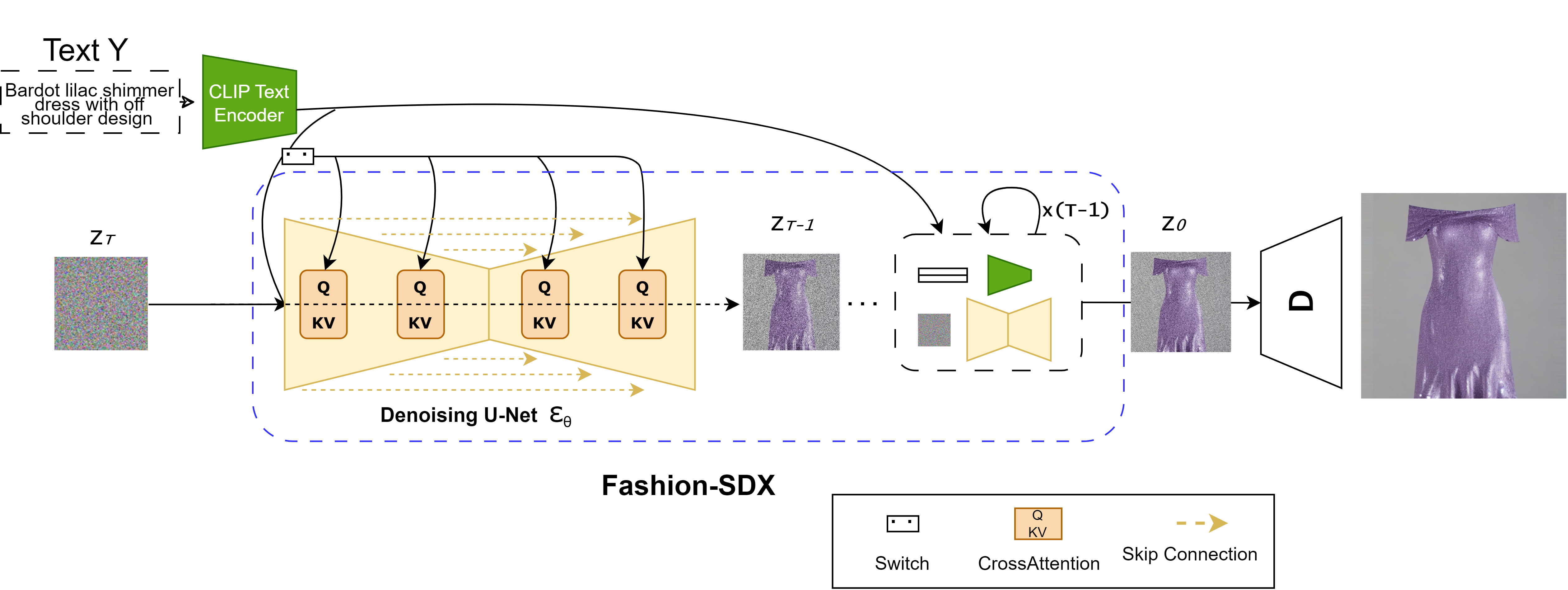}
    \caption{Overview of the Architecture of the proposed FashionSDX , a fashion-centric fine-tuned stable diffusion model}
    \label{fig:archipipeline}
\end{figure*}

In this work, we explore and define a new task of Novel Fashion Garment generation using Stable Diffusion guided and conditioned by multimodal inputs like text prompts, sketches, and texture. For this, we implement a novel pipeline combining Stable Diffusion Low-Rank Adaption (LoRA) and ControlNet. We created two separate pipelines for this task, The first pipeline Fig.\ref{fig:archipipeline} takes the stable diffusion model from Hugging Face and fine-tunes it on our dataset using LoRA method. In the second pipeline, we first train a ControlNet on sketches as an added condition and then use our first pipeline with the ControlNet, to give us a stable diffusion model capable of generating high-quality fashion garments using just text and a sketch.  This will help in manifesting the concept of a dress a fashion designer has in their mind in the form of an image which they can then later use as inspiration. The text prompt provides high-level semantic information about the desired garment, such as its type, color, and style. The sketch provides a detailed outline of the garment's shape and structure. By combining these two input forms, we aim to create a model that can generate detailed, realistic images of garments that closely match the user's specifications.
This work builds upon previous research in text-to-image synthesis, sketch-based image generation, and ControlNet and extends it to the specific domain of fashion garment generation. We believe that this approach has the potential to significantly improve the quality and realism of generated garment images, and could have wide-ranging applications in the fashion industry, like automated design.

For this task, we used two existing virtual Try-On Datasets, Dress Code \cite{b7} and VITON-HD \cite{b8} and we also made use of the Multimodal Dress Code and Multimodal VITON-HD which were extended by Morelli et al\cite{b1,b2} to add conditioning factors like fine grain textual descriptions of the garments, model sketch, model pose, segmentation maps, etc. We further extended these datasets to extract and add the sketches of the garments in the dataset. We then converted this into a hugging face dataset format. We also propose a new evaluation metric that gives us a measure of how structurally and visually similar the generated dresses are to the given sketches. \\

\textbf{Contributions}:
To sum up, our contributions are:
\begin{itemize}
    \item We implement a novel pipeline of a Stable Diffusion LoRA with ControlNet trained on the virtual try-on benchmark datasets like Dress Code and VITON-HD, which to our best knowledge has never been implemented before in the fashion domain.
    \item We introduce a novel fashion-centric generative model to help fashion designers based on latent diffusion models conditioned on multimodal prompts like texts, sketches/textures. 
    We extended the datasets further by extracting sketches using adaptive thresholding and then creating a new hugging face dataset.
    \item A novel evaluation metric to indicate how much reliance the input sketches have on the final model output and how structurally similar the output is to the input sketch.

\end{itemize}

\section{Related Work}

\textbf{Text-to-Image Synthesis:}.
\\
In the context of Text-to-Image synthesis, GANs \cite{b9,b10,b11,b12} have been a popular choice for image synthesis due to their ability to generate realistic images. However, despite their success, GANs have limitations, including mode collapse and training instability, which have led researchers to explore alternative generative models. In most cases now, Diffusion models \cite{b13,b14,b6} are considered to be the state of the art when talking of text-guided image generation. 
In the fashion domain, only a handful of works have been done to this date, earlier works used GANs. One very popular approach is FICE: Text-Conditioned Fashion Image-Editing With Guided GAN Inversion by Pernus et al \cite{b16} incorporates a latent code regularization technique to enhance the GAN inversion process. This technique leverages CLIP textual embeddings \cite{b17} to guide the image editing procedure. In another study, Zhu et al.\cite{b15} introduced a GAN-based method that utilizes both textual descriptions and semantic layouts to generate the final output. On the other hand, Jiang et al. \cite{b18} put forward an architecture that synthesizes full-body images by mapping textual descriptions of clothing items to one-hot vectors. However, this approach restricts the expressive capacity of the conditioning signal.

\textbf{Diffusion Models}
Diffusion models, which were initially introduced by Sohl-Dickstein et al. (2015)\cite{b3} are considered to be fundamental in the line of research in the field of image synthesis \cite{b21,b5,b22,b20}. These models work by transforming a simple prior distribution into a complex data distribution through a sequence of small diffusion steps, which are typically modeled as Gaussian noise. The Denoising Diffusion Probabilistic Models (DDPMs) \cite{b5} introduced a significant improvement over the original diffusion models. DDPMs utilize a denoising process where the noise added at each step is gradually removed, resulting in a sample from the data distribution. This approach has been shown to generate high-quality samples and applied to various domains, including image synthesis. Nichol et al. \cite{b20} built further on the work in \cite{b5} by learning the variance parameter of the reverse diffusion process and producing output with fewer forward passes while maintaining sample quality.\\

While most of the works until now were focused on working in the pixel space,  Rombach et al. \cite{b6} proposed a variant working in the latent space of a pre-trained autoencoder, enabling higher computational efficiency. This allowed for the models to become more flexible and expressive in modeling data distribution.
Song, Meng, and Ermon (2021) introduced the Denoising Diffusion Implicit Models (DDIM)\cite{b23}, a new sampling algorithm that significantly improves the quality of generated samples. DDIM uses a denoising autoencoder to model the data distribution and employs an implicit modeling approach to generate new samples.
Schölkopf et al. (2021) introduced the Stable Diffusion Models\cite{b6}, which further improve the stability and performance of diffusion models. Diffusion models' impact has quickly become disruptive in a variety of tasks, including text-to-image synthesis \cite{b14}, image-to-image translation, image editing, and inpainting \cite{b1,b2}. In the domain of fashion domain, works have majorly been done on virtual try-on tasks, Bhunia et al. \cite{b24} addressed the problem of pose-guided human generation by creating a texture diffusion block based on cross-attention and conditioned on multi-scale texture patterns from the encoded source image. Two notable works, that contributed the most are Multimodal Garment Designer and LaDI-VTON by Davide Morelli et al\cite{b1,b2}. They developed state-of-the-art virtual try-on tasks using a Stable Diffusion textual inversion-enhanced model. In \cite{b1,b2} Morelli et al. used a stable diffusion inpainting pipeline with a modified architecture to factor multiple modalities like sketch, human pose, etc.\\

\textbf{Multimodal Image Generation with diffusion models.}\\
In multimodal image generation, several works have leveraged diffusion models to incorporate various conditioning signals, such as strokes and sketches, to guide the image synthesis process.
The work by Chenlin Meng et al., titled "SDEdit: Guided image synthesis and editing with stochastic differential equations" \cite{b25}, presents an approach that adds noise to a stroke-based input and applies the reverse stochastic differential equation to synthesize images. This method does not require additional training, making it a flexible and efficient solution for guided image synthesis.

Other recent works have explored the use of sketches as additional conditioning signals. For instance, the work by Shin-I Cheng et al.\cite{b26}, proposes to concatenate the sketches with the model input. This approach allows the model to directly utilize the sketch information during the image synthesis process.
Similarly, the work by Andrey Voynov et al., titled "Sketch-Guided Text-to-Image Diffusion Models" \cite{b27}, proposes to train an MLP-based edge predictor to map latent features to spatial maps. This approach allows the model to generate images that closely follow the provided sketches, resulting in more accurate and realistic images.
These works demonstrate the versatility and potential of diffusion models in the field of multimodal image generation. They provide various methods to incorporate different conditioning signals, such as strokes and sketches, to guide the image synthesis process, resulting in high-quality and realistic images. In contrast to all the works mentioned above, we propose a fashion-centric diffusion model that generates high-quality fashion garments by directly exploiting the semantic descriptions of the garment and garment sketches.

\section{Methodology}
\begin{figure*}
    \centering
    \includegraphics[width=1.0\textwidth]{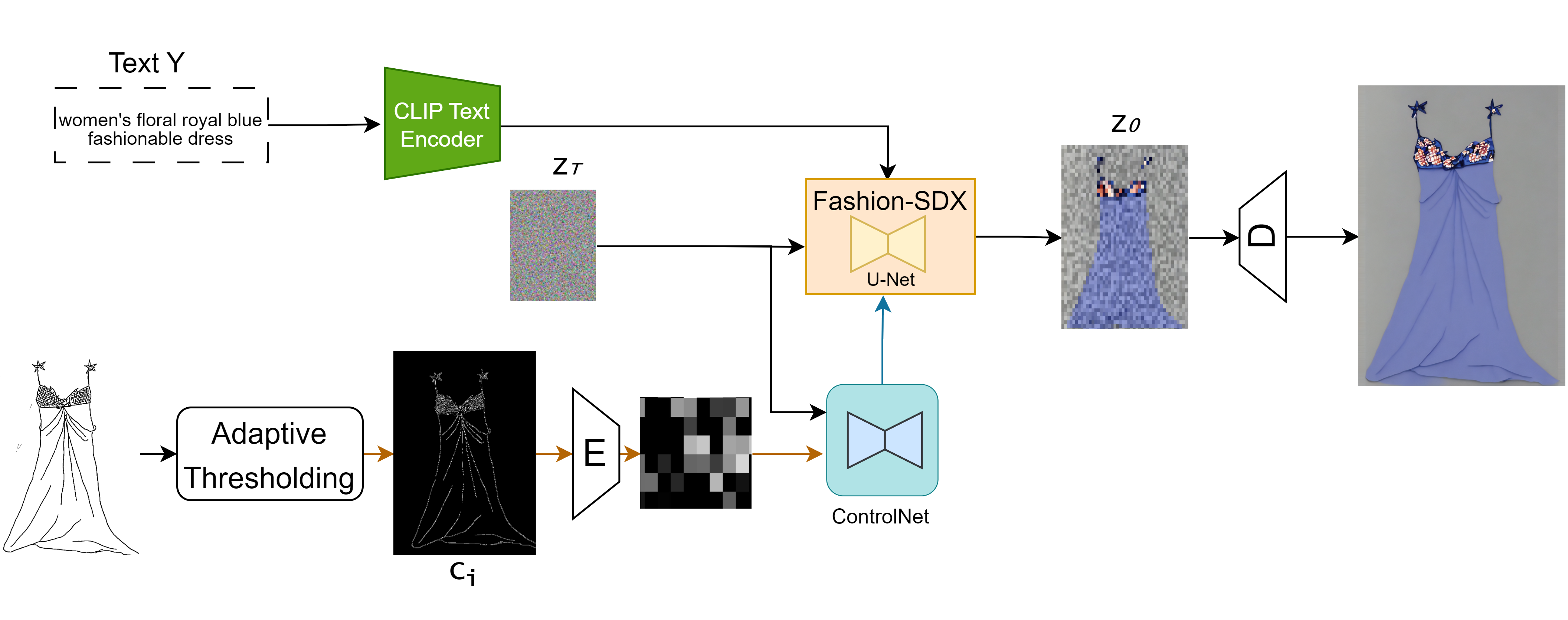}
    \caption{Overview of our Pipeline for Integrating sketches to guide the diffusion process of FashionSDX model}
    \label{fig:pipeline1}
\end{figure*}

In this section, we propose a novel method for the generation of fashion garments with
images conditioned on text and a sketch of the garment. Specifically, given the garment image \begin{math} I \in \mathbb {R}^{W \times H \times 3}\end{math}, a textual description Y of a garment, and a sketch \begin{math} S \in \mathbb {R}^{W \times H \times 3}\end{math}, we want to generate a new image \begin{math} I' \in \mathbb {R}^{W \times H \times 3}\end{math} that generates the picture of the garment in accordance to the given text and sketch inputs. 
We propose two separate pipelines for this task, 1) LoRA Fine-tune model: Fashion-SDX presented in Fig. \ref{fig:archipipeline} and 2) LoRA + ControlNet Fashion- SDX presented in Fig.\ref{fig:pipeline1}.  The first pipeline is a text-to-fashion image model, which is capable of generating diverse and aesthetic results conditioned just on text prompt. In our second pipeline, we incorporated sketches as an added conditioning. We trained our own ControlNet conditioned on sketches from our extended dataset, we added this trained ControlNet to our fine-tuned Fashion-SDX. Now in the following section, we will discuss all the preliminaries that are essential for our study.

\subsection{\textbf{Preliminaries}}

\textbf{Stable Diffusion:}
\\
Stable Diffusion is a latent diffusion model, which consists of an Autoencoder \( \mathcal{A} \) with an Encoder \( \mathcal{E} \), Decoder \( \mathcal{D} \). It also has a text-time conditioning denoising U-Net \( \mathcal{E_{\theta}} \), a CLIP text-encoder \( \mathcal{T}_{E} \)  which takes the input text prompt and converts and maps a sequence of input tokens to a sequence of latent text-embeddings. The Image \begin{math} I \in \mathbb {R}^{3 * W * H}\end{math} is compressed by the Encoder \( \mathcal{E} \) into a latent space of lower dimension \begin{math} \in \mathbb {R}^{4 * w * h}\end{math}, where w = W / 8 and h = H / 8. The Decoder \( \mathcal{D} \) inverts and converts the latent variables back to pixel space.
 The training of the denoising network U-Net \( \mathcal{E_{\theta}} \) is performed minimizing the loss function defined as follows: \\

\begin{math}
  \mathrm{L} = \mathbb{ E}_{ \mathcal{E}(I),Y,\epsilon\sim \mathcal{N}(0,1),t }[\,\lvert\lvert\mathcal{E} - \mathcal{E_{\theta}}({\gamma},{\psi})\rvert\rvert_2^2\, ] ,        
\end{math}   \hfill - (1)\\

Where $t$ is diffusion timestep and $\gamma = z_t$, $z_t$ is the fully noised image or encoded image \(\mathcal{E}(I)\) which is produced by gradually adding Gaussian Noise \(\mathcal{N}(0,1)\) and \({\psi} = [t;\mathcal{T}_{E}(Y)]\), where \( \mathcal{T}_{E}(Y)\) is the encoded text prompt by the CLIP Text Encoder.
In the above equation, ${\gamma}$ is the spatial input of the convolutional network (U-Net) and 
\({\psi} = [t;\mathcal{T}_{E}(Y)]\) is the attention conditioning input.
\\
We use the Standard Stable Diffusion Pipeline, for our task with the aim to generate an Image I of the garment alone without any artifacts or human model wearing the dress. To do this we fine-tune the Stable Diffusion model on our dataset to learn the style of the output we expect.
\\
\textbf{Low Rank Adaptation:} \\
Low-Rank Adaptation(LoRA) \cite{b28} is a parameter-efficient training technique initially used for fine-tuning Large Language Models (LLMs). LoRA reduces the number of trainable parameters by learning the pairs of rank-decomposition matrix while freezing the original weights. These pairs of rank decomposition matrices are called update matrices and are added to existing weights. This technique can also be used in other Large models like Stable Diffusion.
\begin{figure}
    \centering
    \includegraphics[width=0.5\linewidth]{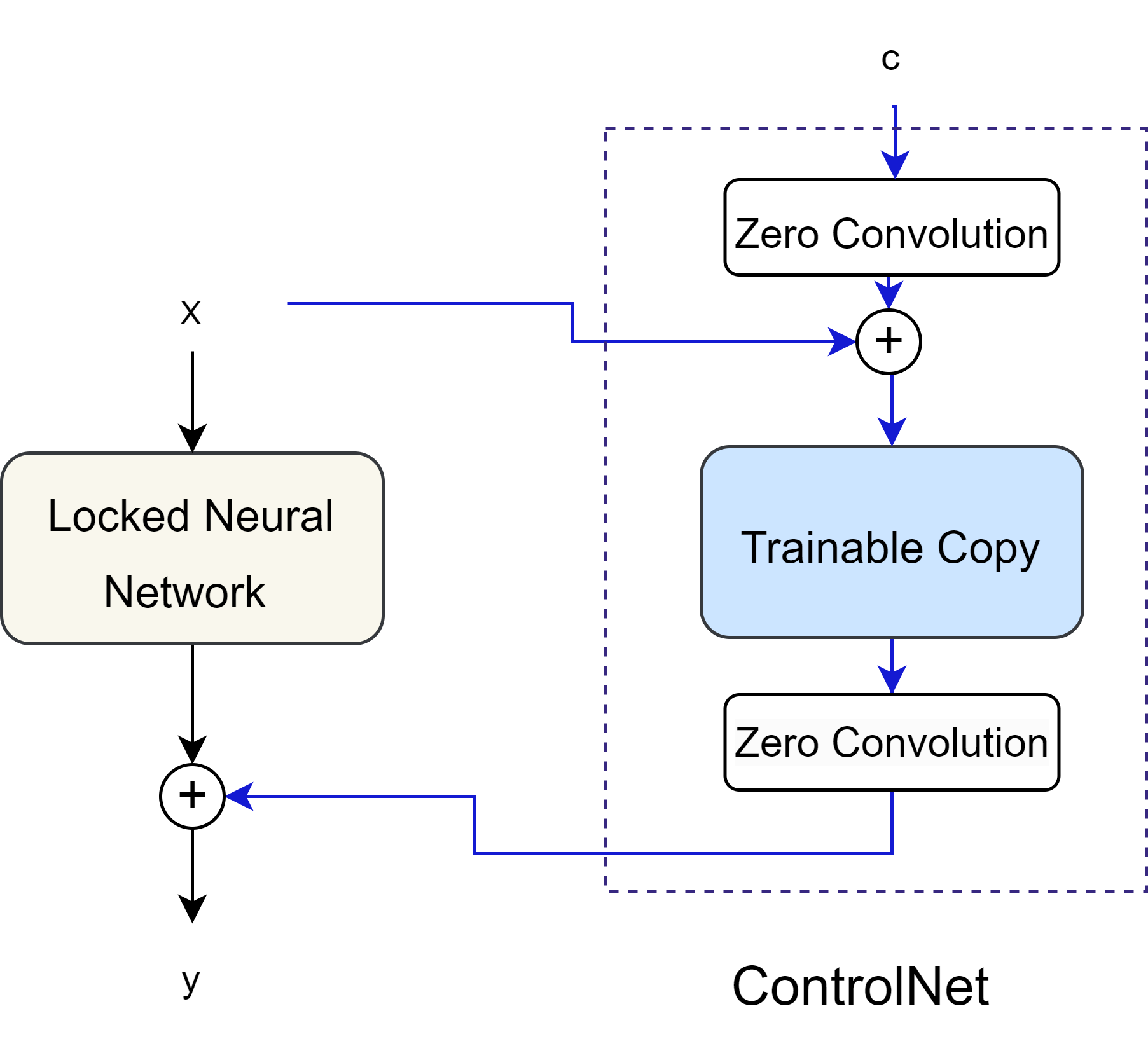}
    \caption{Basic Architecture of ControlNet Network}
    \label{fig:controlnet}
\end{figure}

~\begin{math}
 ~ \mathrm{W'} = \mathrm{W} + \Delta\mathrm{W} \hfill - (2)
~\end{math}\\
In the LoRA fine-tuning process we are only training the $\Delta\mathrm{W}$ in eq. (2). Where we can further decompose $\Delta\mathrm{W}$ into low rank matrices $\Delta\mathrm{W} = AB^T$ where \begin{math} A \in \mathit {R}^{\mathtt{n}  \mathtt{x} \mathtt{d} }\end{math}  and \begin{math} B \in \mathit {R}^{\mathtt{m}  \mathtt{x} \mathtt{d} }\end{math} where $d << n$. So we can just fine-tune A and B matrices instead of W. The resultant trained weights are much smaller in size only $\sim (2-200 Mb)$.
In the case of stable diffusion models, only Q,K,V i.e the attention layers are enough to fine-tune the entire model. As can be seen from Figure \ref{fig:archipipeline}.\\
\textbf{ControlNet:} \\
ControlNet\cite{b29} is a neural network designed to control pretrained large diffusion models and support any additional input conditioning. The controlnet makes a clone or copies the weights of the diffusion model into a "Traininable Copy" and a "Locked Copy", here only the trainable copy is made to learn the task-specific conditioning by training on the task-specified dataset. The trainable and locked copies of the network are linked by Zero-convolutions. The training is fast and comparable to time fine-tuning time because zero convolutions do not add any new noise to the deep features. The ControlNet architecture is shown in Figure. \ref{fig:controlnet}.

\textbf{CLIP:} \\
CLIP is a transformer-based vision-language model\cite{b17} which consists of a vision transformer for building image embeddings and a text transformer for generating text embedding. It aligns the text and visual inputs into a common embedding space.

The visual encoder and text encoder extract these feature representations respectively, \begin{math} \mathit{V_E(I)} \in \mathbb {R}^{d}\end{math} has a dimensionality of 'd' for an input Image $\mathit{I}$,
the text encoder \begin{math} \mathit{T_E(E_L(Y))} \in \mathbb {R}^{d}\end{math} has a dimensionality of 'd', for a text input of Y. In the above, $\mathit{E_L}$ is the embedding lookup that maps every tokenized word in $\mathit{Y}$ to the token embedding space $\mathcal{W}$, and 'd' is the size of CLIP embedding space.

\subsection{\textbf{Fashion-centric Image Synthesis}}
In our task, we aim to generate accurate results from text prompts that are not just realistic and coherent to the text prompt but also in line with what the designers have in mind. To tackle this task of realistic fashion garment synthesis, we took the stable diffusion pipeline by hugging face's diffusers library and fine-tuned the model to learn the style of the in-shop garments. In our pipeline, we start from a clothing garment in our dataset C, which is extracted from the CLIP visual encoder, the text prompt is then projected to the latent space using the CLIP text encoder. We then gradually add noise to the image latent vectors for some given timestep and predict the added noise to the image latent. A reconstruction loss is calculated between the predicted noise and the original noise. Finally, the diffusion model parameters of the U-net are optimized with respect to this loss using gradient descent. The pre-trained text and image encoders are kept frozen. 
The LoRA finetuning significantly helps in training the model fast and helps guide the generation process to properly resemble the style of in-shop garments in our training dataset.\\
\textbf{Sketch Driven Generation:} \\
It is very hard to fully describe the clothing garment you want just based on a textual description due to many factors like ambiguity of natural language and other complexities such as technical terminology. The text might fail to convey certain intricate details such as the size, shape, and texture of the desired garment. Therefore, we proposed the idea of incorporating sketches and texture information of the dresses with our model to further facilitate the fashion generation process. To do this, we incorporated ControlNet into our pipeline.\
\subsection{\textbf{Training}}
We followed the script \cite{b30} given by huggingface to finetune and train a stable diffusion model, in our case we chose to use the Low-rank adaptation. The proposed denoising U-net is trained to predict the noise being stochastically added to z according to Equation given below:\\
\begin{math}
  \mathrm{L} = \mathbb{ E}_{ \mathcal{E}(I),Y,\epsilon\sim \mathcal{N}(0,1),t }[\,\lvert\lvert\mathcal{E} - \mathcal{E_{\theta}}({\gamma},{\psi})\rvert\rvert_2^2\, ] ,        
\end{math}\\

For training on ControlNet, first, we need to decide which stable diffusion model we need to train and which part of the Stable Diffusion we want to train. Usually, all weights inside the ControlNet are copied from the stable diffusion model we select and no layer is trained from scratch, just like fine-tuning. We can also decide which part of the ControlNet and Stable diffusion we want to train. From Fig. \ref{fig:controlnet} we can see that ControlNet has a trainable copy and a locked copy, we can decide to unlock the bottom layers of our locked neural network, this will train a few layers of the original stable diffusion model. We decided to use this functionality for our specific task since our dataset is diverse enough and we think this might help in domain-specific tasks such as Fashion Generation. We also need to decide on a condition to train ControlNet on, We chose to train the network on two separate conditions one sketch and sketch + texture. Our pipeline for generating images conditioned on sketch and text is shown in Figure.\ref{fig:pipeline1}.

\section{Experimental Setup}
\subsection{\textbf{Dataset}}

\textbf{Datasets:} \\ 
\begin{table}[]
\centering
\caption{Table 1. Comparison of the publically available datasets for fashion }
\begin{adjustbox}{max width=0.5\textwidth}
\begin{tabular}{|l|l|l|l|}
\hline
Dataset           & No. of Images & No. of Unique Texts & No. of Sketches\\\hline
DeepFashion       & 44,096        & 10,253              & -              \\\hline
Be Your Own Prada & 78,979        & 3,972               & -              \\\hline
VITON-HD (Ours)   & 13,679        & 5,143               & 13,679         \\\hline
DressCode (Ours)  & 53,792        & 25,596              & 53,596         \\\hline
\end{tabular}
\end{adjustbox}
\label{tab:table1}

\end{table}
There are currently very limited datasets available for fashion generation tasks, most of the datasets that are available lack the conditioning information like text descriptions or sketches. For our task which requires us to finetune a latent diffusion model to generate fashion-centric images using multimodal conditions like text descriptions and sketches. That's why we opted to go for the famous Fashion Try-On Datasets which were further extended by Morelli et al \cite{b1}, known as Dress Code Multimodal and VITON-HD Multimodal, the authors incorporated textual sentences and morphed garment sketches for the virtual try-on tasks. Both these datasets include images of resolution 1024 x 768, each with an image pair of a garment and the corresponding model wearing the garment. Dress Code Multimodal contains 53,792 images of garments split into 3 classes, namely upper body, lower body, and dresses. This dataset contained two prompt files:-\\
Fine-Grained Text Prompts: This prompts file contains manually annotated prompts for 26,400 garments out of 53,792 garments in the dataset.\\
Coarse-Grained Texts Prompts: For the remaining garments were automatically annotated by a fine-tuned OpenCLIP ViT-B32 model pre-trained on LAION5B by Morelli et al.\cite{b1}.\\
We converted both of these text files into two JSON Lines files, and randomly selected and merged the above-mentioned annotations in a single file with mapping to the image paths in the train and test folders. We then split the JSON files into test metadata files and train metadata files. We then use huggingface's Image Folder library to create a dataset in a column format, where the first column has Images loaded in PIL Format and the second column has their respective text prompt in a string format.
A similar, strategy was used to annotate all the garments of the VITON-HD dataset.
For our task, in both of the datasets, we only take the in-shop garments and their corresponding textual descriptions from the multimodal datasets and we remove all other multimodal information such as human pose, human key points, and morphed garment sketches. We then extracted the sketch information of each garment using our custom framework.\\
\textbf{Incorporating Sketches:} \\
Sketch augmentation to our pipeline is essential to provide the users the added functionality to have control over the generated outputs according to their expectations because there might be cases where discrepancies occur when relying solely on text prompts. The resulting pipeline would allow the users to procure outputs that are more coherent, comprehensive, and accurate representations of their expectations of the garments, which in most cases would lead to a better quality of the generated results. Although the multimodal datasets we chose for our tasks have sketches they are not sketches of the in-shop garments but sketches of the garments morphed on the human model's body. 
Instead, we chose to extract sketches of all the garments in the two datasets using the Canny Edge detection algorithm and our own Adaptive Thresholding. 
We had to create a mapping of each sketch to its respective image. Due to memory issues, we had to extract sketches on the fly for every single image and create the mapping at the same time instead of extracting all sketches first. After mapping each sketch to its reference image, we created two new datasets which have an added column with the sketch, this will be our added conditioning.
There are very limited fashion datasets that are publicly available and the ones that are available don't have in-shop garment images which is crucial in our task. Out of these, there are only two such datasets that have corresponding text captions associated with each garment. These datasets are an extension of DeepFashion Dataset\cite{b31}.
We show in Table \ref{tab:table1}. the distribution of the words and texts in these datasets compared to the ones we used. We can see that the dataset we have selected is much more verbose than the competitors and also has sketches associated with each garment.
\subsection{\textbf{Training and Inference}}
\begin{figure}
    \centering
    \includegraphics[width=1.0\linewidth]{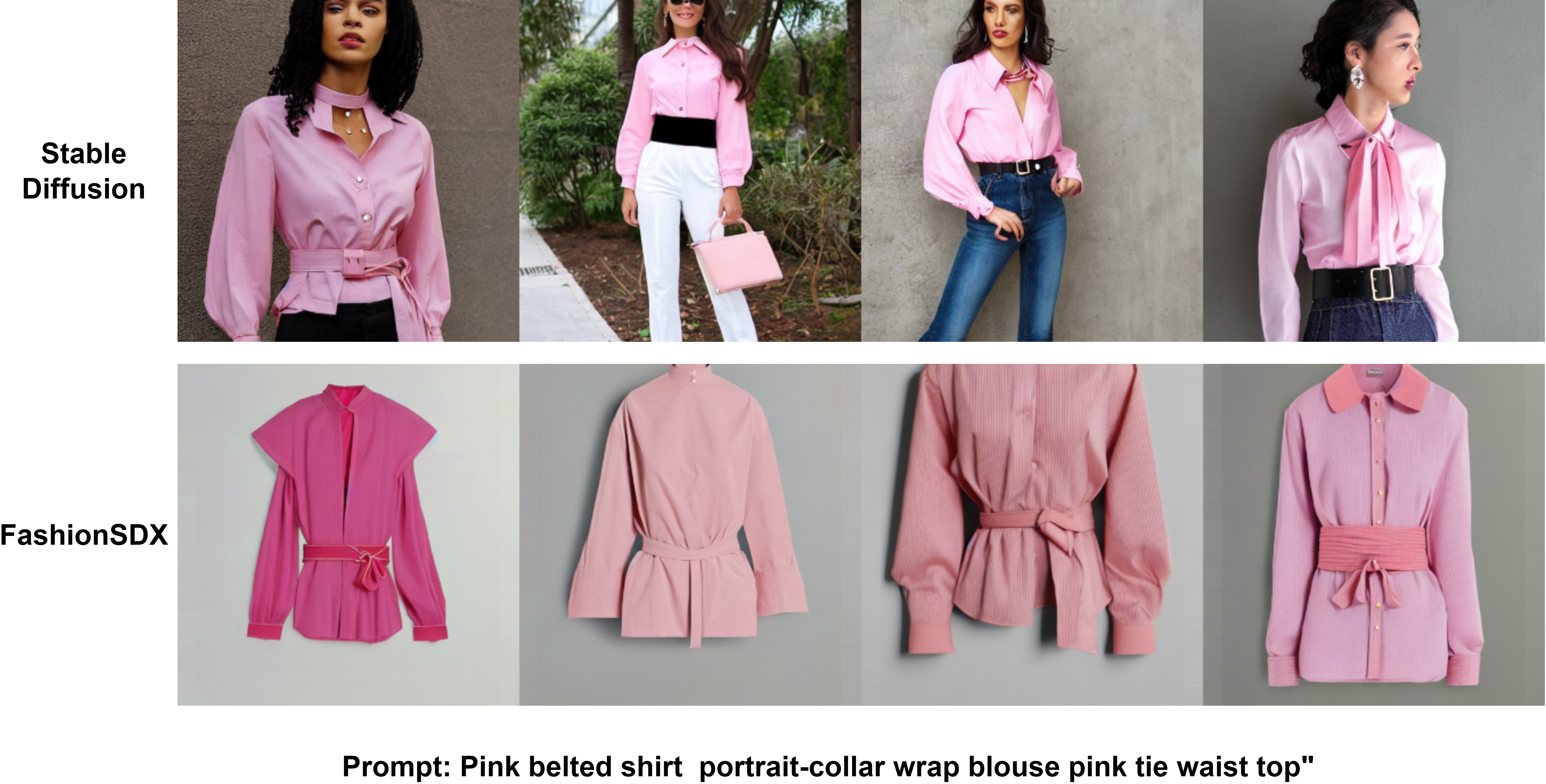}
    \caption{Comparison of Stable Diffusion v/s our trained model for the same prompt.}
    \label{fig:pink}
\end{figure}
All the models were trained on the same system on a single Quadro RTX 6000 (24 Gb VRAM),  Intel CPU XEON Gold 5222 with RAM of 24 GB.
We trained two models via Low-rank adaptation on Dress Code and VITON-HD datasets, for 181k Steps (15 Epochs) and 58k Steps (10 Epochs) for the Dress Code and VITON-HD datasets respectively, at the batch size of 4 and batch size of 2 respectively.  The learning rate in the above two models was \begin{math} {10}^{-4}\end{math} with a linear learning rate warmup for the first 500 steps. We make use of the accelerate library \cite{b32} and script provided by huggingface to speed up the training process, where we opted for mixed precision training 'fp16' due to limited computation resources and to speed up the training process. The model selection was a crucial process, we chose runwayml's stable-diffusion-v1-5 training on the "laion-aesthetics v2 5+" \cite{b33} dataset. The reason why we chose this model is that it performed really well in synthesizing realistic and aesthetic outputs and the input resolution was $512 \times512$ with 10\% text drop rate which improved classifier-free guidance. \\
For training the ControlNet pipeline, we trained two models on VITON-HD, and Dress Code on our extended datasets. For VITON-HD, we trained the model for 5 epochs (18k steps) at a batch size of 4, and for Dress Code, we trained our model for 5 epochs (24k Steps) at a batch size of 4. The learning rate for all ControlNet models is \begin{math}
    {10}^{-5}
\end{math} and we used a constant learning rate scheduler. \\

During Inference, we used PNDMScheduler\cite{b34} and DDIM Scheduler \cite{b23} with the number of inference steps set to 30 and the classifier-free guidance scale $\alpha$ set to 7.5. During, the sketch-guided pipeline inference, we used the PNDMScheduler with a sketch conditioning scale set to 0.6.
For comparison to other baselines, we compared our results with the vanilla Stable Diffusion model from Huggingface's runwayml's stable diffusion-v1-5 and compared all our trained models to results generated from the vanilla stable diffusion.\\

\section{Experimental Evaluation}
\subsection{\textbf{Evaluation Metrics}}
To evaluate our results for generated fashion images on the basis of realism and coherence to the style of our input dataset, we chose to employ the following evaluation metrics: the Frechet Inception Distance (FID) \cite{b35}, FID-CLIP\cite{b36} which uses CLIP feature space for computing FID and the Kernel Inception Distance (KID)\cite{b37}. We used the implementation for both these metrics proposed by Parmar et al. in \cite{b38}. 
We also used Structural Similarity, SSIM for measuring the image quality of generated outputs using sketch. We do this by first extracting a sketch using edge detection from output results and then comparing them using SSIM index with the input sketches.

To verify the adherence of our generated results to the text we used CLIP Score using the implementation provided in the TorchMetrics library, using the OpenAI clip-vit-base-patch16, and the implementation provided by HuggingFace.
Since our results could not be entirely evaluated by qualitative metrics, we chose to do a qualitative survey using our generated results. We asked fashion designer students studying at the London College of Fashion to provide us with sample prompts and gave them 50 images generated by our model and Vanilla Stable Diffusion then asked them to tell us which results they preferred when comparing the two for the 50 sample images. 
Table \ref{tab:table3} shows the percentage of times the users preferred our model compared to results generated by Vanilla Stable Diffusion.

\textbf{Sketch Similarity:}
We also propose, a novel evaluation metric to estimate the adherence of the generated images to the conditioning constraints in our case the sketches. To compute the similarity we first extract the sketch from the generated images using adaptive thresholding and then compute the Mean Squared Error between the extracted sketch and the given sketch's per pixel distance between the two inputs. 
$$
SSim(I_S, \tilde{I_S}) = {MSE}({Sketch(I_S), Sketch(\tilde{I_S})})
$$
\begin{figure}
    \centering
    \includegraphics[width=1\linewidth]{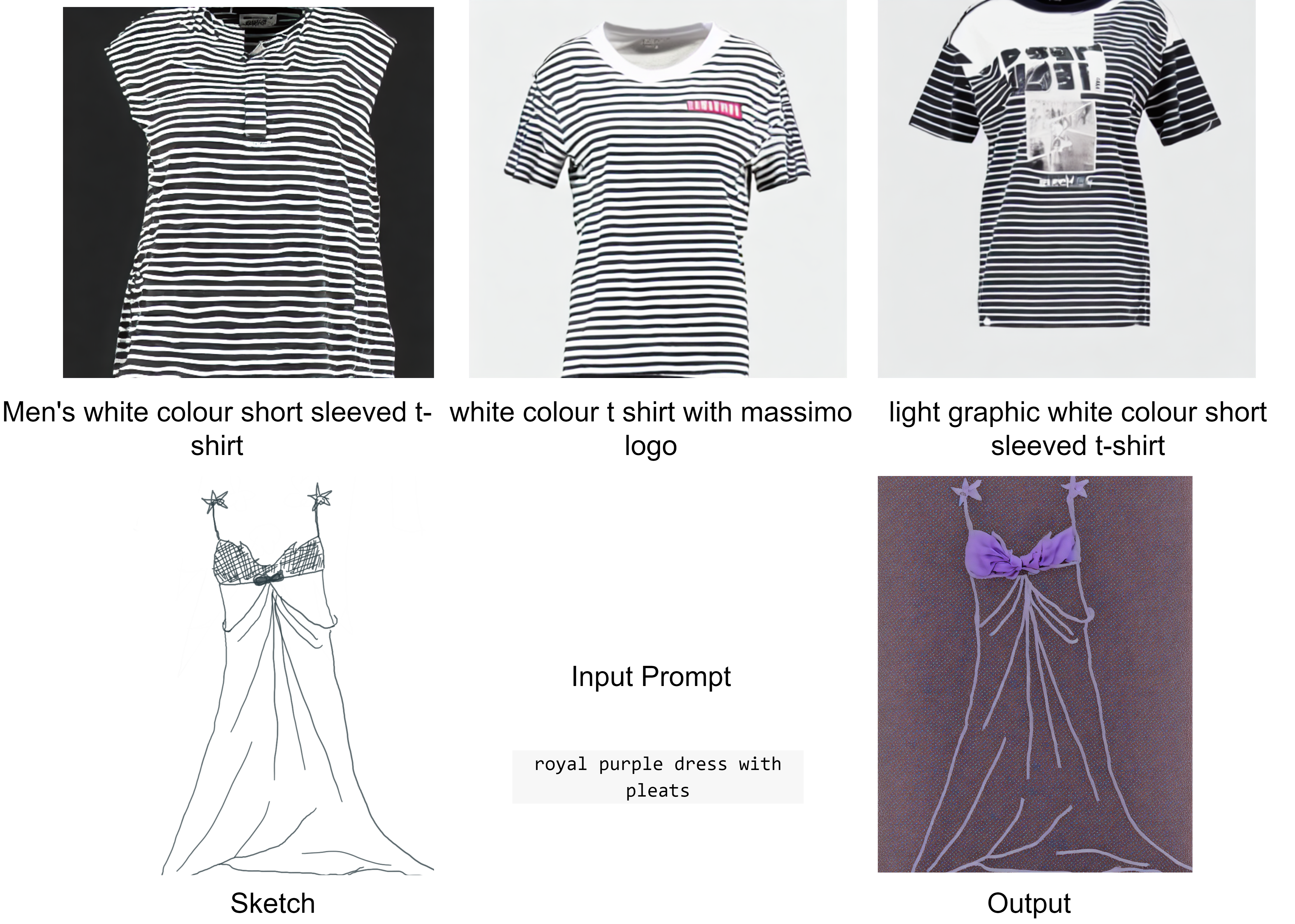}
    \caption{This Figure Displays some of our failure cases, a) First row displays VITON-HD model's failure to comprehend white colour in prompts. The first image also fails to generate male clothes due to the disproportionate distribution of male and female clothes in VITON-HD Dataset.
    b) The Second Row shows how the ControlNet gave results without training, it changed the background according to the prompt but failed to make the garments in accordance to sketch.}
    \label{fig:failcases}
\end{figure}
\subsection{\textbf{Experiments}}
Since we trained multiple models on our two datasets, we tried to experiment with different hyperparameters and different training techniques. In this section, we will discuss all the different approaches we took throughout this project before finalizing our final models. The details of our experimentation for the VITON-HD training are shown in our wandb profile \href{https://wandb.ai/muridayo/text2image-fine-tune}{here}.
We initially, trained our first fine-tuning model without LoRA, it suffered from forgetting and took a long time to train, due to scarce computational resources we had to stop the training prematurely and switch to LoRA fine-tuning for our models. 
On testing, we found that the model trained on VITON-HD for 151k steps overfit easily, it had learned the style of the trainset too well. There, were certain, patterns that kept showing up in the results, and one more prominent and problematic issue was that most of the results even after we specifically mentioned against in prompts were women's wear. This could be because of two reasons, the proportion of female garments in the VITON-HD was much much higher than the men's garments, and the number of samples in the VITON-HD was not enough for us to perform fine-tuning for 151k steps, so we ended up overtraining the model. One more interesting thing we noticed was that when we gave the prompts for any garment with white colour in it, our VITON fine-tuned model always gave black and white stripes, this could be the fault of the prompt given during training. To solve, these issues we trained the models for much fewer steps around 52k, and the results were of much better quality. \\

\begin{table*}[]
\begin{tabular}{|l|l|llll|llll|}
\hline
Model                    & Resolution &                             & Dress                       & Code                        &            &                             & VITON-HD                    &                             &            \\ \hline
                         &            & FID                         & FID-CLIP                    & KID                         & CLIP-Score & FID                         & FID-CLIP                    & KID                         & CLIP-Score \\ \hline
Stable Diffusion         & 512 x 512  & \multicolumn{1}{l|}{84.434} & \multicolumn{1}{l|}{32.993} & \multicolumn{1}{l|}{0.0522} & 32.1053    & \multicolumn{1}{l|}{84.434} & \multicolumn{1}{l|}{32.993} & \multicolumn{1}{l|}{0.0522} & 32.1053    \\ \hline
FashionSDX (Ours)        & 512 x 512  & \multicolumn{1}{l|}{23.804} & \multicolumn{1}{l|}{13.344} & \multicolumn{1}{l|}{0.0082} & 32.1498    & \multicolumn{1}{l|}{42.657} & \multicolumn{1}{l|}{15.257} & \multicolumn{1}{l|}{0.0180} & 31.0188    \\ \hline
Fashion SDX + ControlNet (Ours) & 512 x 512  & \multicolumn{1}{l|}{20.512} & \multicolumn{1}{l|}{16.397} & \multicolumn{1}{l|}{0.0110} & 32.5812    & \multicolumn{1}{l|}{34.659} & \multicolumn{1}{l|}{19.453} & \multicolumn{1}{l|}{0.0288} & 31.5135    \\ \hline
\end{tabular}

\caption{Quantitative results of FashionSD-X trained on Dress Code and VITON-HD and compared to results from a pre-trained Stable Diffusion Model.}
\label{tab:table2}
\end{table*}
When experimenting with the ControlNet pipeline, initially we attempted to add the pre-trained model directly with our FashionSD-X model in the pipeline, The results were quite interesting, the results given by this pipeline failed to comprehend sketch boundaries and instead of creating a garment that adheres to the sketch, it gave the sketch back as it is but changed the background to match the text prompt. 
These findings motivated us to train our own ControlNet model, conditioned on the sketches extracted by us in our extended dataset. The details for training the ControlNet are shown in our wandb profile \href{https://wandb.ai/muridayo/controlnet}{here}.

\begin{table}[]
\centering
\caption{Results of a user study conducted by us, where the users were shown a random 50 images generated by stable diffusion and our models. The users were asked to pick either Stable Diffusion or Our model based on the Realism of the generated garments and coherence to the text prompts. The results here are the percentage of times the users preferred our results compared to stable diffusion}
\begin{adjustbox}{max width=0.5\textwidth}
\begin{tabular}{|lllll|}
\hline
Models                  & Inputs &        & Realism & Prompt Coherence \\
                        & Text   & Sketch & SD      & SD               \\ \hline
FashionSD-X (VITON-HD)  & Y      & N      & 51.45\% & 62.69\%          \\
                        & Y      & Y      & 64.58\% & 67.2\%           \\ \hline
FashionSD-X(Dress Code) & Y      & N      & 84.84\% & 82.1\%           \\
                        & Y      & Y      & 86.26\% & 82.66\%          \\ \hline
\end{tabular}
\label{tab:table3}
\end{adjustbox}
\end{table}

\begin{table}[]
\caption{The table compares the adherence of the outputs to the input sketch using a new metric Sketch Similarity (the lower the value the closer the results to the sketch) and SSIM (the closer the value to 1 the closer the results to the input sketch).}
\begin{adjustbox}{max width=0.5\textwidth}
\begin{tabular}{|l|ll|ll|}
\hline
Model            & Sketch                          & Similarity & SSIM                            &          \\ \hline
                 & \multicolumn{1}{l|}{Dress Code} & VITON-HD   & \multicolumn{1}{l|}{Dress Code} & VITON-HD \\ \hline
Stable Diffusion & \multicolumn{1}{l|}{4525}       & 4421       & \multicolumn{1}{l|}{0.70}       & 0.67     \\ \hline
FashionSD-X      & \multicolumn{1}{l|}{3581}       & 3896       & \multicolumn{1}{l|}{0.78}       & 0.68     \\ \hline
\end{tabular}

\label{tab:table4}
\end{adjustbox}
\end{table}

Fig.\ref{fig:failcases} Illustrates a few of the findings we got via experimentation, on the VITON-HD Dataset and the ControlNet Pipeline.
We learned a lot from our above-mentioned mistakes and then fine-tuned two separate pipelines Fashion-SDX and ControlNet + FashionSD-X trained on Dress Code and VITON-HD successfully. 

\section{Results}
We test our proposed pipeline against the pre-trained stable diffusion model using quantitative and qualitative metrics as described in the previous section. Table II compares our models FashionSD-X and FashionSD-X + ControlNet with the Stable Diffusion results on the basis of FID Score, FID-CLIP score, KID, and CLIP Score. 
To evaluate our results, we generated outputs using our test set's prompts and extracted sketches using thresholding of the garments in our test and then fed all the prompts and sketches to our two pipelines. 
As can be noticed, from the results the FID Score for all the models is quite high, this can be attributed to the background of the results being drastically different. The results would be much closer to zero if we had segmented the clothes and pasted them on 512 $\times$ 512 white background. Nonetheless, Both our models outperform Stable Diffusion based on Realism - FID, FID-CLIP, and KID Scores. The Closer the FID Score and KID scores are to zero the less distance there is between our generated images and sample images in our test dataset in terms of realism.
The CLIP-Score of our model on many occasions struggled to outperform Stable Diffusion, but the final results when computed for the entire test dataset were found to be quite comparable for the Dress Code dataset. The CLIP score is highly dependent on the model used for the encoding process, the discrepancy found in our results even though visually superior to Stable Diffusion can be possibly attributed to its failure to understand the nuances of our specific fashion domain.

We notice from Table \ref{tab:table2} that our models trained on VITON-HD, outperformed Stable Diffusion results in terms of FID and KID scores but the CLIP-Score was lower than Stable Diffusion. The results for our models trained on the Dress Code dataset were far superior to the ones from the VITON-HD dataset. The reason for this is possibly due to the smaller size of samples in VITON-HD (13k images) with less diverse data than the Dress Code data (54k images). We showed some of the failure cases where the model trained on VITON-HD failed to generate "White t-shirt" in fig. \ref{fig:failcases}. 
\begin{figure}
    \centering
    \includegraphics[width=1\linewidth]{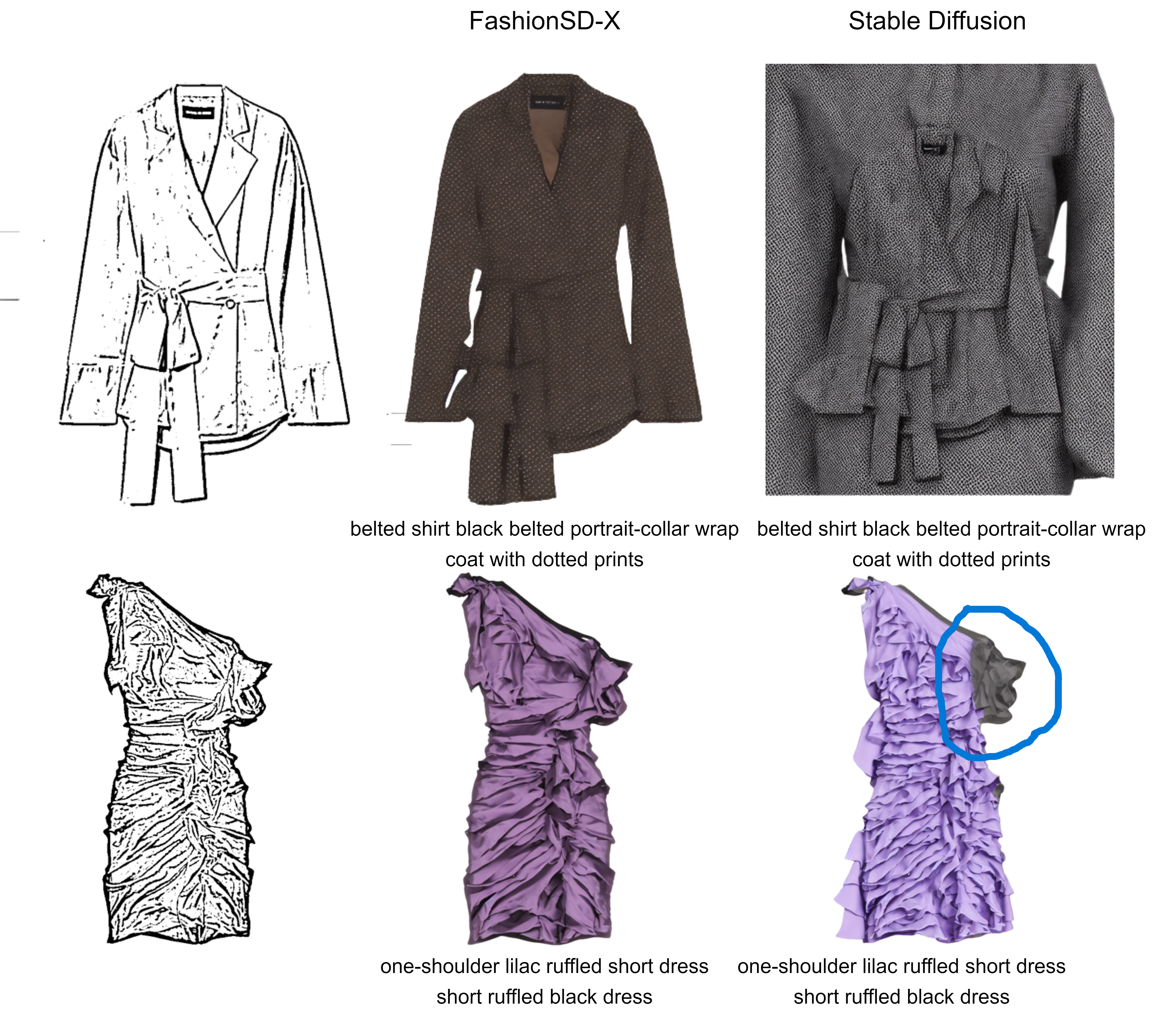}
    \caption{Sample Images generated on Dress Code dataset, demonstrating how our model's performance is superior in terms of quality of results and coherence to prompts.}
    \label{fig:sd vs fash}
\end{figure}

We used a new metric called Sketch Similarity to compare how close the outputs are to input sketches, the results in Table \ref{tab:table4} show the Sketch Similarity which is the MSE between the input sketches and the sketches extracted from the output dresses using Canny Edge Detection. We observe that our ControlNet pipeline with FashionSD-X performs much better than Stable Diffusion on both datasets in both metrics Sketch Similarity and Structural Similarity Index. We observed from the stable diffusion's results conditioned on sketch that most of the time the results bleed outside of the sketch boundary, and there's a morphed human present in the results. Fig \ref{fig:sd vs fash} showcases the sample output for the same sketch and prompts how Stable Diffusion's outputs bleed outside the sketch boundary. \\

For the qualitative results, highlighted by Fig \ref{fig:showcase}, Fig \ref{fig:pink}, Fig \ref{fig:sd vs fash} and Table \ref{tab:table4} we can observe based on the generated results that the quality of fashion garments generated by our model is far superior to Stable Diffusion in terms of realism and quality.
We also conducted a mini user-study, to validate our results in accordance with professional and human judgment.
We collected responses from 10 Fashion Design students at the London College of Fashion and asked them to select their preference between a pair of images one generated by our model and the other by stable diffusion, we did this for over 50 randomly sampled images from our generated images.
The results are presented in Table \ref{tab:table2} which shows the percentage of times our model was preferred by the users in the study, We observe that our model trained on Dress Code performed phenomenally conditioned on both sketch and text, the users preferred our model's output over stable diffusion in almost all cases. We also noticed that the results generated by our model on the VITON-HD dataset were quite inferior compared to Dress Code. The users preferred stable diffusion's output almost 48\% times over our model trained on VITON-HD.

\begin{figure}
    \centering
    \includegraphics[width=1\linewidth]{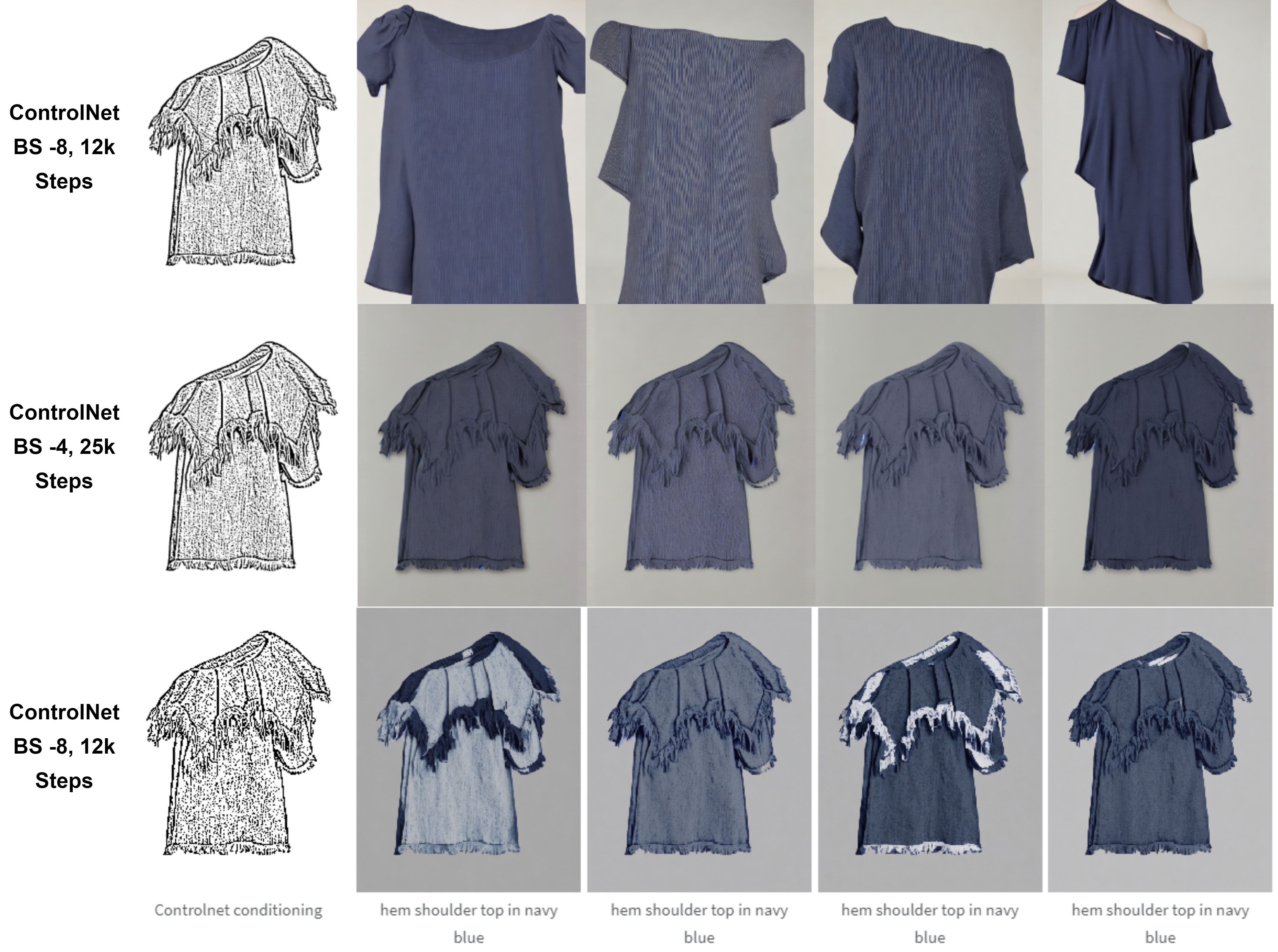}
    \caption{Abalation results for ControlNet Training.}
    \label{fig:contf}
\end{figure}

We experimented with batch size and the number of training steps during the training of the ControlNet part, We trained three different models a) Batch Size = 4, No. of Training steps = 25k, b) Batch Size = 4, No. of Training Steps = 18k and c) Batch Size = 8 , Steps = 12k. We also performed validation on the two models to see their performance qualitatively. The model c) had the worst performance when compared to the other two. The qualitative results for this can be seen in Fig. \ref{fig:contf}
The results were the best from the model that was trained for 25k steps, the model also had a high structural similarity index when compared to the other two controlnet models. \\

During inference, we experimented with a number of Noise Schedulers, When experimenting with DDIM Noise Scheduler we noticed something rather interesting this scheduler was always giving images that kept getting flagged as NSFW images, and a black image was displayed, even though when we disabled the safety checker the results the results came out as black images. This was not experienced, with any other noise scheduler we experimented on. This is something that requires further work. \\

\section{Conclusion}
This research  explores the multifaceted landscape of diffusion models and their application to fashion design. We provide an exhaustive investigation of  diffusion model's  potential, challenges, and contributions to fashion. We propose a novel pipeline, for generating fashion garments conditioned on sketch and text. We delve into the  interplay between visual and textual cues, different datasets, training strategies, and evaluation metrics. The exploration unfolds a comprehensive understanding of diffusion models, not only enhancing the theoretical understanding of generative models but also bridging the gap between artificial intelligence and creative design in the dynamic fashion industry. The results showcase one of the most significant works in this field in doing a fashion designer's job and helping them in the designing of innovative and creative garments. Our work stands as a testament to the flexibility and robustness of diffusion models, positioning them as vital tools in modern design and artificial intelligence research.

\vspace{12pt}

\end{document}